\pdfoutput=1
\documentclass[english]{lni}

\usepackage{color}
\usepackage{paralist}
\usepackage{amsmath}
\usepackage{pifont}
\usepackage{booktabs}
\usepackage{subcaption}
\usepackage{multirow}
\usepackage{csquotes}

\newcommand{\cmark}{\ding{51}}
\newcommand{\xmark}{\ding{55}}

\usepackage{array}
\usepackage{makecell}

\begin{document}
\title[On the State of German (Abstractive) Text Summarization]{On the State of German (Abstractive) Text Summarization}
\author[Dennis Aumiller \and Jing Fan \and Michael Gertz]
{Dennis Aumiller\footnote{Heidelberg University, Institute of Computer Science, Im Neuenheimer Feld 205, 69120 Heidelberg, Germany \email{aumiller@informatik.uni-heidelberg.de}} \and
Jing Fan\footnote{Heidelberg University, Institute of Computer Science, Im Neuenheimer Feld 205, 69120 Heidelberg, Germany \email{j.fan@stud.uni-heidelberg.de}} \and
Michael Gertz\footnote{Heidelberg University, Institute of Computer Science, Im Neuenheimer Feld 205, 69120 Heidelberg, Germany \email{gertz@informatik.uni-heidelberg.de}}}
\startpage{1} 
\editor{B. K{\"o}nig-Ries et al.}
\booktitle{Datenbanksysteme f{\"u}r Business, Technologie und Web (BTW 2023)}
\yearofpublication{2023}
\maketitle


\begin{abstract}
With recent advancements in the area of Natural Language Processing, the focus is slowly shifting from a purely English-centric view towards more language-specific solutions, including German.
Especially practical for businesses to analyze their growing amount of textual data are text summarization systems, which transform long input documents into compressed and more digestible summary texts.
In this work, we assess the particular landscape of German abstractive text summarization and investigate the reasons why practically useful solutions for abstractive text summarization are still absent in industry.\\
Our focus is two-fold, analyzing a) training resources, and b) publicly available summarization systems.
We are able to show that popular existing datasets exhibit crucial flaws in their assumptions about the original sources, which frequently leads to detrimental effects on system generalization and evaluation biases.
We confirm that for the most popular training dataset, MLSUM, over 50\% of the training set is unsuitable for abstractive summarization purposes.
Furthermore, available systems frequently fail to compare to simple baselines, and ignore more effective and efficient extractive summarization approaches.
We attribute poor evaluation quality to a variety of different factors, which are investigated in more detail in this work:
A lack of qualitative (and diverse) gold data considered for training, understudied (and untreated) positional biases in some of the existing datasets, and the lack of easily accessible and streamlined pre-processing strategies or analysis tools.
We therefore provide a comprehensive assessment of available models on the cleaned versions of datasets, and find that this can lead to a reduction of more than 20 ROUGE-1 points during evaluation.
As a cautious reminder for future work, we also highlight the problems of solely relying on $n$-gram based scoring methods by presenting particularly problematic failure cases.
The code for dataset filtering and reproducing results can be found online: \url{https://github.com/dennlinger/summaries}

\end{abstract}

\section{Introduction}

Libraries simplifying the access to pre-trained neural models have greatly pushed the recent advancement of state-of-the-art performance in many tasks~\cite{wolf-etal-2020-transformers}.
However, with the general absence of non-English resources, one of the prevalent challenges in the Natural Language Processing (NLP) community is the extension of approaches to other languages beyond English.
Subsequently, evaluation quality and consistency is even harder to maintain in setups, where high-quality gold data is scarce.
This can lead to unintended consequences during the interpretation of model performance and generalization capabilities beyond narrow domain-specific use cases.

A sub-task of the NLP community that deserves particular attention is text summarization. The focus here is to produce an abridged version of an input text that accurately \emph{summarizes} the key points of the original text. Such systems offer an immediate benefit in times with ever-increasing amounts of textual information, and allow users to quickly grasp the contents of even complex documents.
In particular, we differentiate between various sub-tasks of text summarization: \emph{extractive} systems provide summaries by simply copying text snippets from the original input, which is efficient to compute, but comes at the cost of lower textual fluency. On the other hand, \emph{abstractive} summarization systems may introduce new phrases, or even full sentences, which are not present in the original document. This potentially increases a summary's fluency and conciseness over extractive methods. Abstractive text summarization systems are generally built upon the more recent development of sequence-to-sequence neural models~\cite{sutskever-etal-2014-sequence, bahdanau-etal-2015-neural}, which come with an exploding computational cost.

Particularly for (abstractive) summarization, the previously mentioned issues of data scarcity for non-English methods are further worsened by a lack of diverse (and readily available) evaluation metrics.
Most works rely entirely on $n$-gram-based analysis of system summaries, such as ROUGE~\cite{lin-2004-rouge}, which cannot accurately judge the truthfulness of a generated summary, i.e., how accurately the original text's factual statements are represented in the generated summary. Only few works extend their evaluation to human result inspections, given its higher cost.
However, there are several critical assumptions that --even under basic premises-- are exposed as oftentimes insufficient for a comprehensive analysis~\cite{sparck-jones-1997-summarising, ter-hoeve-etal-2020-what}. Examples are the focus on singular target summaries, ignoring the subjective nature of differing viewpoints of annotators, as well as the focus on particularly prominent sentences in the first few paragraphs of reference articles~\cite{zhu-etal-2021-leveraging}.

\noindent In this work, we focus on German abstractive summarization systems and set out to investigate, reproduce, and evaluate summarization systems. In conjunction to a model-centric view of summarization, we further review the existing training resources for German, including their particular domain and data curation processes.

\begin{enumerate}
	\item We find that in particular automatically created and multilingual resources suffer from insufficient pre-processing, potentially due to the absence of a native speaker during the curation process.
	\item News documents seem to be overly represented in trained systems, potentially due to a popularity bias in English summarization datasets for news resources.
	\item Baseline scores are heavily affected by data biases in test sets of prominent datasets.
\end{enumerate}

Upon conducting a qualitative analysis of outputs from publicly available models, we further find that most systems fall severely short of the expected quality in at least one of the following areas:
\begin{enumerate}
	\item Due to positional biases, text snippets may be directly copied from the beginning of the input text, constituting an \emph{extractive} instead of an \emph{abstractive} summary. Especially considering the computational requirements of neural systems being orders of magnitudes greater than simple extractive summarizers, this undermines the quality of neural text generations.
	\item Generated outputs may contain (severe) syntactic errors, to the point of becoming illegible or hard to interpret.
	\item Semantic mistakes introduce factual errors, leading to incorrect conclusions from the summary alone. This problem is exacerbated for longer input documents, where a structured content understanding is necessary to maintain factual consistency.
\end{enumerate}

For data-centric issues, current pipelines are not taking user-specified filtering steps into account; oftentimes, datasets are directly used ``out of the box'', without any further data verification step involved. For this purpose, we extend available summarization-specific filtering steps and provide a simple-to-use and language-agnostic processing library.\\
For model-centric problems, it is near impossible to identify failure cases with existing metrics; costly manual inspection of individual samples would be required. Simultaneously, we work towards expanding the available scores to help facilitate a better understanding of current expectations towards summarization systems.
In the following, we will briefly mention work on automated evaluation of summarization systems, including a comprehensive look at the current landscape of German abstractive summarization; we follow with a formal introduction of our proposed filtering methods for summarization datasets, as well as a list of model-centric checks to consider.
We discuss exhibited quality issues in existing datasets and systems for German summarization, and conclude with a brief outlook for future work.



\section{Related Work}

We establish an extensive overview of currently available training resources for German summarization systems and survey the landscape of trained models, with a particular focus on publicly available methods.\\
Aside from this, we further reiterate some of the common pitfalls in evaluating summarization systems, which will become particularly relevant during the experiments in this work.

\subsection{German Data Sources for Summarization}
In our experiments, we focus on seven different datasets across a variety of domains. To our knowledge, these cover all of the publicly available sources used for training German systems.

\paragraph{MLSUM~\cite{scialom-etal-2020-mlsum}}
This multilingual dataset was presented as one of the first efforts in making larger-scale training sets available for multiple languages that also include German as a language. MLSUM is constructed by extracting news articles and associated summary sections as generation targets.
We use the German subset in this work, which is by far the most popular dataset used for training and evaluating resources in German, based on our survey.
Despite its popularity, issues in the quality of samples have gone unnoticed until early 2022, when Philip May~\cite{may-2022-anomalies} was the first to report on problems with fully extractive summaries, an aspect we will analyze in more detail later.

\paragraph{MassiveSumm~\cite{varab-schluter-2021-massivesumm}}
The construction of this particular dataset is similar to MLSUM and focuses on a large number of automatically extracted summaries from news articles in multiple languages.
The authors perform some rudimentary filtering with respect to empty samples and even go as far as avoiding similar issues to MLSUM by removing what they call ``ellipsoid summaries'', i.e., fully extractive summaries that appear at the beginning of the reference text.
While the quality of the samples is comparatively low due to the automated extraction process, this corpus is by far the largest considered, with around 480{,}000 samples, and has the potential to improve existing training setups with its sheer number of samples.

\paragraph{Swisstext~\cite{frefel-etal-2020-german, frefel-2020-summarization}}
In contrast to the --generally shorter-- news articles available in MLSUM, the Swisstext dataset provides longer-form summaries based on German Wikipedia pages, which has been later extended to the GeWiki corpus~\cite{frefel-2020-summarization}.
For the construction, the central argument is that the introductionary paragraph serves as a ``summary'' of the remaining article text. The provided dataset comes with a training portion and a private test set, meaning no ground truth summaries are available for the test samples.
A multilingual variant of this idea, the XWikis corpus, was introduced shortly after~\cite{perez-beltrachini-lapata-2021-models}. While the XWikis corpus contains more samples per language, including for German, monolingual data is not readily available for download. Adding the fact that German summarization works primarily deal with the Swisstext dataset, we choose the latter for our experiments.

\paragraph{Klexikon~\cite{aumiller-gertz-2022-klexikon}}
Another Wikipedia-related resource, but with different target summaries. Instead of utilizing a page's introductionary paragraph, the authors align articles from a simplified children's encyclopedia (Klexikon) on the same topic.
Consequently, this dataset has much longer summary lengths but covers a much smaller subset of only around 3{,}000 samples.
Given the secondary focus on simplification in the target summaries, this corpus requires a considerably higher level of abstractive reformulations during the generation.

\paragraph{WikiLingua~\cite{ladhak-etal-2020-wikilingua}}
As the third multilingual resource, summaries in this corpus are extracted from the WikiHow platform.
Here, Ladhak et al.~\cite{ladhak-etal-2020-wikilingua} consider short instruction summaries of individual steps in WikiHow guides and align those with the referenced paragraphs.
The general tone of the dataset is rather informal and is in a more imperative style in comparison to other data sources.
To align non-English samples, associated images are used to identify paragraphs occurring in different languages. Importantly, this means that for German articles, frequently only some of the article's paragraphs are actually contained in the dataset.

\paragraph{LegalSum~\cite{glaser-etal-2021-summarization}}
Another area benefiting enormously from high-quality summaries is the legal domain. LegalSum is the first German resource providing summaries of around 100{,}000 court rulings.
On average, these samples require the highest amount of compression across evaluated datasets.

\paragraph{EUR-Lex-Sum~\cite{aumiller-etal-2022-eur}}
As a secondary resource for legal texts, Aumiller et al.~present a multilingual corpus based on EU-level legal acts, semi-aligned across languages. The corpus is considerably smaller than LegalSum, with only about 1{,}900 German documents available.
While the EUR-Lex-Sum corpus has extremely long documents, it also presents a more challenging summary generation with the longest average summary length across all considered corpora (generally between 600-800 words).
Importantly, summaries are also written by human expert annotators and therefore present a much higher-quality standard for summaries compared to some of the other datasets.

\paragraph{Further Resources}
In addition to these datasets, we are aware of at least two more news-related resources. One is used in experiments by Nitsche~\cite{nitsche-2019-towards}, where data was supplied by the German Press Agency, but no public record of it exists. The second corpus is hinted at online by users on Huggingface's platform\footnote{A news corpus with ca.~400{,}000 articles is indicated here: \url{https://huggingface.co/Einmalumdiewelt/PegasusXSUM\_GNAD/discussions/1\#6308eb5037556c4ab03258df}, last accessed: 2023-01-14}. \\
For clinical summarization, Liang et al.~\cite{liang-etal-2022-fine} present a resource of about 11{,}000 radiology reports; given the sensitive nature of the data, no publicly available version exists as of now. 
We are also aware of a secondary source of the WikiLingua dataset by GEM\footnote{\url{https://gem-benchmark.com/data_cards/wiki_lingua}, last accessed: 2023-01-14}, which provides additional samples, as well as a pre-split validation and test section not provided in the original German subset. In preliminary experiments, we found that $>99.89\%$ of the data were valid samples for the GEM source. Most problematic is the automatic combination of paragraphs into one summary, which can cause disjoint reference texts or summaries.\\
Finally, all of the discussed corpora so far are types of single document summarization resources, where a summary is extracted from a \emph{singular text only}. Datasets for training summarization systems that consider multiple source texts exist at smaller scales, but require further manual adjustment for acquisition~\cite{benikova-etal-2016-bridging,zopf-2018-auto}. More recent experiments with neural models utilizing the latter corpus have been conducted by Johner et al.~\cite{johner-etal-2021-error}.

\subsection{German Summarization Systems}

\begin{table*}[ht]
	\setlength{\tabcolsep}{3.5pt}
	\centering
	\hspace*{-0.5em}
	\begin{tabular}{l|cccc|cc}
		\textbf{Model} & \textbf{Training data} & \textbf{Test Set} & \textbf{Evaluation} & \textbf{Filtering} & \textbf{Public} & \textbf{Reprod.} \\
		\hline
		mrm8488/bert2bert\footnotemark & MLSUM & MLSUM & ROUGE & None & \cmark & \cmark\\
		ml6team/mt5-small\footnotemark & MLSUM & MLSUM & ROUGE & Length & \cmark & \xmark \\
		T-Systems/mt5-small\footnotemark & \makecell{CNN/DailyMail,\\MLSUM, XSum,\\Swisstext} & MLSUM & ROUGE & \makecell{Length \&\\ Overlap} & \cmark & \xmark \\
		Shahm/t5-small\footnotemark & MLSUM & MLSUM & ROUGE & None & \cmark & \xmark \\
		T5-base\footnotemark & \textbf{?} & \textbf{?} & ROUGE & \textbf{?} & \cmark & \xmark \\
		german-t5\footnotemark & Swisstext & MLSUM & ROUGE & \textbf{?} & \xmark & \xmark \\
		BERT-Copy~\cite{aksenov-etal-2020-abstractive} & Swisstext & Swisstext & \makecell{ROUGE \&\\manual} & \textbf{?} & \cmark & \textbf{?} \\
		Transformer~\cite{parida-motlicek-2019-swisstext} & \makecell{Swisstext \&\\CommonCrawl} & Swisstext & \makecell{ROUGE \&\\manual} & None & \xmark & \xmark \\
		Fact-Encoder~\cite{venzin-etal-2019-swisstext} & Swisstext & Swisstext & \makecell{ROUGE \&\\manual} & None & \xmark & \xmark \\
		Pointer-Gen~\cite{fecht-etal-2019-swisstext} & Swisstext & Swisstext & \makecell{ROUGE \&\\manual} & \textbf{?} & \xmark & \xmark \\
		Enc-Dec~\cite{glaser-etal-2021-summarization} & LegalSum & LegalSum & ROUGE & \textbf{?} & \cmark & \textbf{?}\\
		bert2bert~\cite{liang-etal-2022-fine} & Radiology & Radiology & \makecell{ROUGE \&\\manual} & \textbf{?} & \xmark & \xmark \\
	\end{tabular}
	\caption{List of German abstractive summarization systems. We detail their known properties from training recipes or papers. If we have access to models, we denote whether public scores are reproducible within $\pm$0.5 ROUGE points (``\emph{Reprod.}''); \textbf{?} in the reproducibility column indicates models that are available, however, we were unable to successfully run locally.}
	\label{tab:models}
\end{table*}

While we are slowly starting to see a greater diversity in the available training resources for German text summarization, it comes as a small surprise that the availability of trained system is much less diverse.
As will become more apparent in later sections, the primary focus for training systems is a combination of a pre-trained checkpoint and one predominant training resource (``MLSUM'', particularly the German subset).
Below, we elaborate on considered model properties, differentiating between the availability levels of related works and their backgrounds. A summary of known properties can be seen in Table \ref{tab:models}.

\footnotetext[6]{\url{https://hf.co/mrm8488/bert2bert_shared-german-finetuned-summarization}, last accessed: 2022-10-06}
\footnotetext[7]{\url{https://huggingface.co/ml6team/mt5-small-german-finetune-mlsum}, last accessed: 2022-10-06}
\footnotetext[8]{\url{https://huggingface.co/T-Systems-onsite/mt5-small-sum-de-en-v2}, last accessed: 2022-10-06}
\footnotetext[9]{\url{https://huggingface.co/Shahm/t5-small-german}, last accessed: 2022-10-06}
\footnotetext[10]{\url{https://huggingface.co/Einmalumdiewelt/T5-Base\_GNAD}, last accessed: 2022-10-06}
\footnotetext[11]{\url{https://github.com/GermanT5/german-t5-eval}, last accessed: 2022-10-06}

\subsubsection{Publicly Available Systems}
The primary source for available models is the Huggingface Hub\footnote{\url{https://huggingface.co/models}, last accessed: 2023-01-14}, which allows filtering by supported language and appropriate task (in our case summarization). We note that some of the available models are not properly tagged, but spent considerable time to ensure no other models were accidentally ignored.
For users who have uploaded several different versions, we selected the model with the highest self-reported evaluation scores.\\
Given that users on the platform are likely familiar with other services of Huggingface (including their datasets browser), it comes as no surprise that the diversity between models and training setups is low.
The primary choice falls on either mT5~\cite{xue-etal-2021-mt5} or variants of T5~\cite{raffel-etal-2020-exploring}, with some alternatives based on (m)BART~\cite{lewis-etal-2020-bart, liu-etal-2020-multilingual} being consistently outperformed according to self-reported metrics by authors.
In order to train effectively on large quantities on data, most approaches use one of the smaller checkpoints, referring to model variants with fewer parameters.
Outside of the model hub, code repositories exist for the BERT-Copy architecture by Aksenov et al.~\cite{aksenov-etal-2020-abstractive} and Encoder-Decoder models used by Glaser et al.~\cite{glaser-etal-2021-summarization}. However, we were unable to set up inference for custom datasets based on the respective code bases.


\subsubsection{Private Models}
A further selection of models has been published in response to the Swisstext 2019 summarization challenge~\cite{parida-motlicek-2019-swisstext, venzin-etal-2019-swisstext, fecht-etal-2019-swisstext}.
However, neither team has published any associated public repository. Similarly, no models are available from Liang et al.'s work on radiology reports~\cite{liang-etal-2022-fine}.
As the only one of the major cloud providers, Microsoft offers an extractive summarization service through Azure that supports German.\footnote{\url{https://learn.microsoft.com/en-us/azure/cognitive-services/language-service/summarization/language-support}, last accessed: 2022-10-06}
Otherwise, the only commercial solution providing a platform for abstractive summarization also supporting German texts is currently Aleph Alpha.\footnote{\url{https://www.aleph-alpha.com/use-cases/conversion\#trilingual-summary}, last accessed: 2022-10-06}

%
%


\subsection{Evaluation Metrics for Summarization}
As previously mentioned, the de-facto gold standard for evaluating summarization systems is the usage of ROUGE~\cite{lin-2004-rouge}. The authors introduce unigram overlap (ROUGE-1), bigram overlap (ROUGE-2) and the longest common subsequence (ROUGE-L) between system and gold predictions.
The underlying core assumption is based on $n$-gram co-occurrences in the generated text with respect to one or more gold summaries.
The fact that ROUGE can handle several reference samples at the same time is crucial for understanding some of the implications in the later parts of this work:
with several references, variation in wording, e.g., particular expressions, are much easier to compare against than in a single reference summary.
However, despite the theoretical support for multi-labels, few datasets ever provide such costly annotations.\\
In turn, more recently proposed alternatives to ROUGE rely on score computation from a single gold summary only~\cite{ermakova-eta-2019-survey}.
Examples include primarily neural similarity scoring between a generated summary and a gold reference~\cite{sellam-etal-2020-bleurt, zhang-etal-2020-bertscore}.
Ultimately, neural methods are also incredibly expensive to employ for evaluation settings, potentially taking several days to evaluate a single test set~\cite{nan-etal-2021-improving}.
Besides the cost factor, the main issue with such alternative scores is two-fold:
On the one hand, a distinct advantage of simple co-occurrence-based metrics such as ROUGE is the simplicity in transferring it to another language. Even basic extensions, such as stemmers, are readily available in a multitude of languages other than English. Trained metrics, such as BERTScore~\cite{zhang-etal-2020-bertscore} or QAEval~\cite{deutsch-etal-2021-towards}, however, are severely limited in their transferability to other languages, and would require dedicated efforts to port them to German, for example.
On the other hand, recent statistical analyses have shown that when accounting for annotator expertise, correlation can vary significantly~\cite{fabbri-etal-2021-summeval}. When additionally controlling for variance and confidence intervals, correlation with human judgments over ROUGE correlation is only statistically significant in rare cases~\cite{deutsch-etal-2021-statistical}.
A particular investigation on metrics for German summarization was conducted during the second Swisstext challenge~\cite{frefel-etal-2020-german}.
However, submitted resources were only marginally better than ROUGE baselines for judging system quality~\cite{paraschiv-cercel-2020-upb,biesner-etal-2020-hybrid}.
For crowd-sourced evaluation approaches, Iskender et al.~\cite{iskender-etal-2020-best} further elaborate on the importance of survey setups and considerations for expert annotators to ground evaluation results.

\section{Assessing the Quality of Summarization Systems}

\begin{figure}[ht]
	\hspace*{-0.25cm}
	\centering
	\begin{tabular}{|l|l|l|}
		\hline
		\textbf{Issue} & \textbf{Reference} & \textbf{Summary} \\
		\hline
		\textbf{Short text} & \makecell[l]{Wir verwenden Cookies, um unser Angebot\\ für Sie zu verbessern. Mehr Informationen [...].} & -- \\
		\hline
		\multirow{2}{*}{\textbf{Duplicates}} & `Virtuelles Bergsteigen mit dem Project360 [...] & Leben und Kultur in Europa \\
		\cline{2-3}
			& Historische Dokumente: Bilder der Wende [...] & Leben und Kultur in Europa \\
		\hline
		\textbf{Relative Length} & \makecell[l]{Chef-Sprüche: ``Ich sehe meine \\Kinder auch nur im Urlaub.''} & \makecell[l]{Die besten Chef-Sprüche zum \\Thema Überstunden.}\\
		\hline
	\end{tabular}
	\caption{List of faulty summarization samples in the MassiveSumm dataset uncovered by various data checks. Despite checking for unrelated issues, we notice a trend where filtered samples are of especially low semantic quality, too.}
	\label{fig:samples}
\end{figure}

When using existing models for abstractive text summarization, the expectation is that they should work ``as expected'', meaning that a model provides appropriate and correct summaries.
However, in practice, the automated collection of samples may lead to insufficient sample quality or systematic biases in the data. This has further detrimental consequences for models trained on those datasets.\\
In this section, we lay out a series of very basic sanity checks for both data and models, which help to ensure a minimal level of generalization from experimental results. As we will later find, even such basic data assurances lead to a significant reduction of valid samples in available German summarization data.

\subsection{Data-centric Sanity Checks}
\label{sec:metrics}
The best strategy to achieve decent experimental results is ensuring high quality in the training data -- in line with the popular saying ``garbage in, garbage out''.
We present a list of minimal quality checks for individual samples, as well as dataset-wide assurances of data quality.
Most of these measures are fully automated and at most require single hyperparameter settings to filter datasets. \\
Further, suggested data checks are language-independent at their core and can therefore be applied in basic form to any dataset, even beyond German.
This also implies that no further existing tools or libraries for tokenization, etc., are required.

\paragraph{Empty Samples}
The most trivial sanity check is verifying that \emph{both} the reference text and summary are present for all samples. This is simultaneously the most prevalent check implemented by authors of resource papers in our experience.
Even so, several issues can arise for this criterion, primarily revolving around varying definitions of ``emptiness''.
For example, one could also consider a sample as empty if only whitespaces (or whitespace-like symbols, such as \texttt{$\backslash$t}) are present. Extensions are, for example, faulty encodings or only special characters in a text (cf., data audit insights by Kreutzer et al.~\cite{kreutzer-etal-2022-quality}).

\paragraph{Minimum Text Length}
A superset of ``empty samples'', imposing a required minimum text length presents a stricter filtering criterion for sample validity.
Where empty texts are universally to be avoided, hard length requirements are harder to impose, since the appropriate cutoff depends strongly on the dataset domain.
For domain-specific datasets, e.g., the instruction-like texts in the WikiLingua dataset~\cite{ladhak-etal-2020-wikilingua}, having extremely short summaries with only a few characters (and comparatively short references) may make sense.
For summaries stemming from news articles, however, length requirements imposed on the reference might ensure a longer minimum text length for quality control.
\paragraph{Compression Ratio Filtering}
Another key metric used in summarization research is the \emph{Compression Ratio (CR)}, defined as the relation between reference text length and summary length. We follow the definition by Grusky et al.~\cite{grusky-etal-2018-newsroom}: $CR(\text{ref}, \text{summ}) = \frac{\text{len}(\text{ref})}{\text{len}(\text{summ})}$.
For filtering by compression ratio, a significant difference should be ensured by establishing a minimum compression ratio. For our purposes, we argue that a reduction of at least 20\% in the summary length is required, which equals $CR \geq 1.25$. We note that this is not a strict requirement per se and may depend on domain-specific factors.
It can be argued, however, that samples with summaries longer (or equal) than their respective references (i.e., $CR \leq 1.0$) always pose an inadequate sample and must be filtered.\\
Related work sometimes takes a more drastic approach to compression ratio filtering, arguing that extreme content reduction may result in a lossy summary and should therefore also be avoided~\cite{urlana-etal-2022-tesum}. 

\paragraph{Duplicate Removal}
Some lesser-checked property seems to be the existence of duplicates in training data, which is also applicable in more general machine learning settings.
However, given that each sample for summarization comes with \emph{two} distinct texts (the reference and the summary), we can further distinguish between different instances of duplication.
Trivial to consider are instances of what we call \textbf{exact duplicates}, i.e., samples that have the exact same combination of reference and summary appearing as another tuple in the dataset. \\
We can further expand this idea by three more considerations, which we call \textbf{partial duplicates}. These are instances where we find either the reference or summary in other dataset instances. Finally, it could also occur that both summary and reference are duplicated, but across different samples; such instances are also considered partial duplicates and are relatively rare.\\
To understand why duplicates, including partial ones, can be considered harmful as a training resource, we need only look at the potential effects during training or evaluation.
For exact duplicates, no real gain is achieved by including one sample several times in the training data. Worse yet, if we encounter exact duplicates \emph{across different splits}, this can cause active falsification of evaluation results (train-test leakage). While partial duplicates are less severe, we still encourage removal, as they can cause confusion during the learning process: cases where different input texts should generate the same summary hamper generalization of models, and the reverse case of similar input texts generating different summaries conveys unclear learning signals during training.
Finally, we also want to note that partial duplicates can uncover incorrectly aligned samples (cf.~\Cref{fig:samples}).\\
While \emph{spotting} duplicates is fairly straightforward, removing duplicate content is often non-trivial, as there exist several valid strategies for deduplication, leading to differing results. In an attempt to reduce impact on smaller test and validation sets, we adopt a ``additive'' strategy for the remainder of this work. We start with an empty dataset, and iteratively add new samples if and only if neither the reference nor the summary have been previously included.

\paragraph{Sample Inspection}
Even with all of the proposed automated measures, nothing can ensure data quality quite as well as manually inspecting data. All of the previous measures can point to systemic failures in the data collection process, but may ignore more localized quality issues for particular samples.
While a manual analysis step is not feasible at scale, often enough reviewing few samples will already reveal tendencies about the underlying data quality.
We generally differentiate between the following strategies to inspect data samples and their respective up- and downsides:

\begin{enumerate} 
	\item \textbf{Reviewing samples in order:} A linear sequence of samples may reveal particular issues in the consistency of samples, which can be linked to the crawling process. We emphasize that ``linearity'' can follow many particular axes, not just the order in which data is stored. Further possible orderings can be based on available metadata descriptors, such as sortings by timestamps, source or length. In-order samples are most likely to uncover systematic issues, such as incorrect alignment settings that span several samples.
	\item \textbf{Reviewing random samples:} Another popular approach is to shuffle data and randomly select instances for review. This is fairly easy to implement and does not require iterating over the full dataset or sorting operations. Advantages of random reviews are a more holistic coverage of the data distribution, but requires potentially more manual reviews to find systemic failures.
	\item \textbf{Outliers and representative samples:} If data statistics are already known or easy to compute, a more targeted approach is to look for distributional outliers. There are again a variety of metrics that can be considered, with the most obvious being text length and compression ratio of individual samples. Manually reviewing outliers can also sharpen the requirements of expected outputs, e.g., the minimum/maximum length of a summary in relation to the input text. Related are \emph{representative samples}, which constitute instances close to the mean or median of a distribution.
\end{enumerate}


\subsection{Model-centric Checks}

While we have compiled a detailed list of what can be done about checking the data used for summarization systems, it is significantly harder to judge a trained system, especially given that many neural methods can only be treated as black box systems.
But even with a lack of clarity around the original training procedures and model learnings, we can use several probing techniques to estimate the robustness and performance of systems.

\paragraph{Evaluation on Cleaned Test Sets} The standard procedure to evaluate on withheld (but still in-domain) test sets.
While these evaluation approaches may give insights on the overfitting of trained models, such experiments tend to fall short of giving more concrete evidence on the pattern of how summaries are generated.
This is especially crucial if no further manual evaluation is performed.
Testing models on modified or generalized test data can serve as a partial remedy to this, by probing the generalization ability of particular systems.
In combination with the proposed filtering techniques, we suggest the evaluation on cleaned test sets for models that were trained on the unfiltered training set.
The main advantage is that no additional re-training with altered training sets is required, and insights can be acquired from a generally much smaller evaluation set through inference alone.
Further, we hypothesize that intrinsic summarization metrics~\cite{narayan-etal-2018-dont,grusky-etal-2018-newsroom, zhong-etal-2019-closer, bommasani-cardie-2020-intrinsic} applied to \emph{system summaries} can be used as a preliminary gauge for text quality in comparison to the original input.
Especially \emph{abstractiveness} of generated outputs, essentially constituting the number of novel $n$-grams in summaries, could indicate changes in the vocabulary.

\paragraph{Domain Generalization}
An extreme case of the previous point is testing on completely out-of-domain data, which usually means taking test splits of a different dataset.
While this approach can be useful to evaluate general purpose summarization systems, the evaluated models in this work all present rather focused domain-specific summarization systems.
For this reason, we refrain from evaluating performance based on out-of-domain abilities.

\paragraph{Factual Consistency} A rather important argument for summarization especially: facts that are stated in the original reference text must be maintained in the respective summary.
Therefore, models should be measured with respect to their truthfulness, which has been previously attempted with automatic metrics for English summarization systems~\cite{kryscinski-etal-2020-evaluating}, or even implemented as an optimization target for more truthful summaries~\cite{zhu-etal-2021-enhancing}.

\subsection{Extractive Models and Baseline Systems}
\label{sec:extractive}
Given the relatively one-dimensional approach to evaluation, we should at least expect additional context for better interpretability of model scores. In practice though, we rarely find a consistent reporting of baseline scores, if any comparisons are reported at all. To this end, we strive to provide consistent baselines and reporting of such in the context of German abstractive summarization. In addition to the scores, baselines also serve an important purpose by providing a sensible complexity trade-off: Unlike most neural methods, they should be able to generate summaries faster and with fewer parameters than heavy-weight state-of-the-art approaches.
Similar to English works, we therefore fall back on extractive summarization systems, which -- as the name indicates -- simply copy text snippets from the reference to generate a summary.\\
To our knowledge, the only work that has explicitly worked on extractive summarization for German is over 20 years old~\cite{reithinger-etal-2000-summarizing}. This does not imply, however, that there is no dedicated extractive system available. Especially for unsupervised methods, such as TextRank~\cite{mihalcea-tarau-2004-textrank} or LexRank~\cite{erkan-radev-2004-lexrank}, language-specific taggers or lemmatizers can easily be replaced in existing libraries to enable application on German texts as well.
For our experiments, we rely on three variants of baselines, which extract a specified number of sentences from the input text to generate a summary. 
Overall, extractive summaries are guaranteed to ensure a more factually consistent summary, and have high intra-sentence coherence.
On the other hand, these methods cannot be fine-tuned and rely on singular hyperparameters -- the length of the generated summary.
This can still significantly impact the evaluation performance, but does not factor in domain-specific variance in text distribution.
For all systems, we rely on the sentence splitting module by spaCy\footnote{We use the model \texttt{de\_core\_news\_sm} in our experiments.}, unless datasets provide a pre-split sentence format.

\paragraph{Lead-3}
The simplest possible baseline system is \emph{lead-3}, popularized by Nallapati et al.~\cite{nallapati-etal-2017-summarunner} as a simple but strong baseline for news article summarization. Here, the summary is equal to the first three sentences of an input text.
The method works particularly well for news texts, where key information has to be conveyed early on to both inform and catch the interest of a potential reader. The prevalence of this so-called ``lead bias'' differs significantly across different domains.

\paragraph{Lead-$k$}
For other domains, three sentences may underestimate the actual summary length. For this purpose, Aumiller and Gertz~\cite{aumiller-gertz-2022-klexikon} introduce a variant that extends the lead baseline to the $k$ leading sentences, in their particular context the full first paragraph of a Wikipedia page.
Given that in general, datasets do not contain paragraph-level information, the authors later extended this baseline and instead consider an approximate $\hat{k}$ for each sample by using the average compression ratio~\cite{aumiller-etal-2022-eur}:
\begin{align}
	\hat{k} = \frac{\text{len(reference)}}{CR_{\text{avg}}},
\end{align}
where $\text{len(reference)}$ is the number of sentences in the summary, and $CR_\text{avg}$ denotes the average compression ratio across the \emph{training} split of a dataset. 

\paragraph{Modified LexRank (LexRank-ST)}
A more complex baseline that also considers sentences at other positions of the article is a modification of LexRank~\cite{erkan-radev-2004-lexrank}, similarly used by Aumiller et al.~\cite{aumiller-gertz-2022-klexikon, aumiller-etal-2022-eur}.
The key modification lies in exchanging the centrality computation -- which is originally based on pure occurrence counts -- with dense sentence embeddings obtained through sentence-transformers~\cite{reimers-gurevych-2019-sentence,reimers-gurevych-2020-making}.
While the underlying neural model can be of arbitrary complexity, it does not need to be trained further to work in the summarization application.
After scoring individual sentences, the highest-ranking $k$ sentences are selected as the summary; we use the same method for estimating an optimal length $\hat{k}$ as for the lead-$k$ baseline.

Finally, we also point towards oracle extractive summaries as a form of upper-bound for extractive summarization, which can be computed from greedy ROUGE-2 alignments~\cite{nallapati-etal-2017-summarunner,glaser-etal-2021-summarization}. Given that we focus on \emph{abstractive} results in this work, we omit the computation of extractive oracle summaries.

\section{A Sober Look at State-of-the-Art Results}

%

Given the presented set of tools, we now set out to put current models' capabilities into a better context. To this end, we conduct a set of four experiments:
We start by applying the filters introduced in \Cref{sec:metrics} to available German summarization datasets, noting varying size reductions as a result.
To remedy the changes introduced by our filtering, we re-compute a set of strong baselines as updated results for datasets with available validation and test sets.
Further, given the previously uncovered discrepancies in some datasets, we repeat more comprehensive experiments on MLSUM and MassiveSumm across the pre- and post-filtered dataset to highlight the effect of filtering on ROUGE scores.
We are able to show that this change in data quality also significantly impacts the reproducibility of results. Finally, we provide a small case study in which we examine a subset of generated samples that highlight some of the particular model-centric issues.

\subsection{Filtering Datasets}

\begin{displayquote}
	\textbf{Key Finding 1:} German subsets of two popular multilingual resources (MLSUM and MassiveSumm) have extreme data quality issues, affecting \textbf{more than 25\% of samples} across all splits.
\end{displayquote}

Table \ref{tab:filter} presents our findings for filtering the available German summarization datasets; hyperparameters for filters are specified in the table caption. We refrain from imposing any particularly strict filtering metrics, particularly for the length of texts.
Most concerning is the fraction of affected samples in MLSUM, given its popularity as a training resource for many public models.
While a strong lead bias is to be expected due to the domain of these samples being exclusively news articles, the eventual performance of models trained on the unfiltered dataset is severely impacted; a finding that we confirm in subsequent experiments.
Primarily, it indicates that for fully extractive samples, summaries can be generated by directly running an extractive summarization system, and thus obtain similar (or better) quality at a much lower cost.
For MassiveSumm, a large fraction of invalid samples can be attributed to duplicate content; manual inspection reveals that there are frequent generic references or summary texts, such as ``\emph{Read more after logging in!}''. We assume the reason to be a faulty extraction of HTML elements for particular websites.\\
The remaining inspected datasets were affected at a much lower rate; we see several subsets that have only a handful of faulty instances.
Depending on the overall size of the dataset, this implies that evaluation scores will differ less between unfiltered and filtered splits of largely unaffected datasets.

\begin{table}[t]	
	\centering
	\hspace*{-0.65cm}
	\setlength{\tabcolsep}{2.25pt}
	\begin{tabular}{ll|c|cc|c|c|c|ccc|cc}
		& & & \multicolumn{2}{c|}{\textbf{Min Length}} & & \textbf{Min} & \textbf{Fully} & \multicolumn{3}{c|}{\textbf{Duplicates}} & &\\
		\textbf{Dataset} & \textbf{Split} & \textbf{Samples} &\textbf{Ref} & \textbf{Summ} & \textbf{Id} & \textbf{CR} & \textbf{Extr} & \textbf{Exact} & \textbf{Ref} & \textbf{Summ} & \multicolumn{2}{c}{\textbf{Valid Samples}} \\
		\hline
		\hline
		& Train & $220{,}887$ & $0$ & $0$ & $39$ & $30$ & $\mathbf{126{,}204}$ & $31$ & $45$ & $105$ & $94{,}433$ & $(42.75\%)$ \\
		\textbf{MLSUM} & Val & $11{,}394$ & $0$ & $0$ & $0$ & $0$ & $\mathbf{3{,}285}$ & $1$ & $1$ & $5$ & $8{,}102$ & $(71.11\%)$ \\
		& Test & $10{,}701$ & $0$ & $0$ & $0$ & $0$ & $\mathbf{3{,}306}$ & $1$ & $5$ & $2$ & $7{,}387$ & $(69.03\%)$ \\
		\hline
		\textbf{MassiveS} & Train & $478{,}143$ & $253$ & $\mathbf{16{,}294}$ & $0$ & $\mathbf{33{,}959}$ & $0$ & $805$ & $\mathbf{73{,}886}$ & $4{,}882$ & $348{,}064$ & $(72.79\%)$ \\
		\hline
		\textbf{Swisstext} & Train & $100{,}000$ & $0$ & $0$ & $0$ & $0$ & $3$ & $0$ & $0$ & $2$ & $99{,}995$ & $(100.00\%)$ \\
		\hline
		\textbf{WikiLing} & Train & $58{,}341$ & $11$ & $0$ & $0$ & $\mathbf{1{,}435}$ & $0$ & $4$ & $2$ & $52$ & $56{,}837$ & $(97.42\%)$ \\
		\hline
		& Train & $2{,}346$ & $0$ & $0$ & $0$ & $10$ & $0$ & $0$ & $2$ & $0$ & $2{,}334$ & $(99.49\%)$ \\
		\textbf{Klexikon} & Val & $273$ & $0$ & $0$ & $0$ & $1$ & $0$ & $0$ & $0$ & $0$ & $272$ & $(99.63\%)$ \\
		& Test & $274$ & $0$ & $0$ & $0$ & $1$ & $0$ & $0$ & $0$ & $0$ & $273$ & $(99.64\%)$ \\
		\hline
		& Train & $1{,}115$ & $0$ & $0$ & $0$ & $18$ & $0$ & $0$ & $0$ & $0$ & $1{,}097$ & $(98.39\%)$ \\
		\textbf{EUR-Lex} & Val & $187$ & $0$ & $0$ & $0$ & $0$ & $0$ & $0$ & $0$ & $0$ & $187$ & $(100.00\%)$ \\
		& Test & $188$ & $0$ & $0$ & $0$ & $0$ & $0$ & $0$ & $0$ & $0$ & $188$ & $(100.00\%)$ \\
		\hline
		& Train & $79{,}937$ & $0$ & $2$ & $0$ & $12$ & $326$ & $233$ & $95$ & $\mathbf{3{,}106}$ & $76{,}163$ & $(95.28\%)$ \\
		\textbf{LegalSum} & Val & $9{,}992$ & $0$ & $0$ & $0$ & $4$ & $32$ & $14$ & $2$ & $157$ & $9{,}783$ & $(97.91\%)$ \\
		& Test & $9{,}993$ & $0$ & $0$ & $0$ & $7$ & $33$ & $8$ & $1$ & $59$ & $9{,}885$ & $(98.92\%)$ \\
	\end{tabular}
	\centering
	\caption{German text summarization datasets in numbers. Given are the original sample count and breakdown of filtered samples by automated assessment (cf.,~\Cref{sec:metrics}) for all provided splits. We set the \emph{Minimum Length} to 20 characters for summaries and 50 for references, except for WikiLingua, which has limits of 8 and 20 characters, respectively, due to a different domain. \emph{Id} refers to samples with same reference and summary text, \emph{Min CR} ensures references are at least 25\% longer than summaries, and \emph{Fully Extr} identifies consecutive segments that are used as fully extractive summaries. For duplicates, we differentiate between both reference and summary appearing in the corpus (\emph{Exact}), versus partial duplicates where only one of reference (\emph{Ref}) \emph{or} summary (\emph{Summ}) are appearing elsewhere. Numbers in bold highlight issues affecting more than 2\% of the split data.}
	\label{tab:filter}
\end{table}

\subsection{Consistent Results and Baseline Runs}
\begin{displayquote}
	\textbf{Key Finding 2:} Existing evaluation scores are hard (if not outright impossible) to reproduce, even with model weights publicly available.\\
	\textbf{Key Finding 3:} Authors frequently fail to put scores into context, not comparing their own results against baseline methods for further scrutiny.
\end{displayquote}

Another worrying trend we observe in the ``reproducibility'' column of \Cref{tab:models}, is the consistent inability to even approximately reproduce self-reported scores for any of the evaluated models.
In our reproduction attempts, we employed no particular further filtering, and observed scores that were anywhere from 5 points worse to 3 points better than self-reported ones on the test set. Only a singular result was reproducible within 0.5 ROUGE points of the expected results.
In particular, we find that implementation details on filtering steps and other subselection criteria are rarely (if ever) included in the documentation of training procedures. While the usage of so-called ``model cards''~\cite{mitchell-etal-2019-model}, i.e., dedicated documentation pages for particular training results, has improved the availability of at least \emph{some form} of documentation, these descriptions are still insufficient to fully reproduce results.
As a side note, it should also be mentioned that multiple implementations for the ROUGE evaluation metric exist\footnote{e.g., rouge-score (\url{https://pypi.org/project/rouge-score/}, last accessed: 2022-10-06) or pyrouge (\url{https://pypi.org/project/pyrouge/}, last accessed: 2022-10-06)}, which may result in scoring differences by utilizing different text processing tools or implementations.
To ensure reproducibility of our own scores, we mention that scores were computed with help of the \texttt{rouge-score} package, version 0.1.2. We further replaced the default stemming algorithm with the German Cistem stemmer~\cite{weissweiler-fraser-2017-cistem} to provide a reasonable upper-bound of scores and use the provided bootstrap sampler with $n=2000$.

Aside from the lack of reproducible results, we also noted that only few public models report against a set of (consistent) baselines, with the most commonly compared approach being lead-3.
Given that we have also presented a cleaned portion of popular evaluation models, we strive for a more comprehensive comparison of actual results, and investigate resulting implications that were omitted in the original evaluation settings.\\
In our particular setup, we compare against the three mentioned extractive baselines mentioned in \Cref{sec:extractive} and report scores in \Cref{tab:baselines}.
Depending on the dataset, the choice of a baseline can heavily skew the interpretation compared to neural methods. For example, on the Klexikon dataset, using lead-$3$ can lead to a roughly 12-13 point drop in ROUGE-1 scores compared to scores by the lead-$k$ or LexRank-ST baseline.
On the other hand, for lead-heavy and short texts in MLSUM, lead-3 serves as the best baseline method. Our recommendation is therefore to similarly use multiple (different) baseline approaches, resulting in a more defined context for evaluation based on ROUGE scores.
While it may be easier to simply copy results from prior work, we highly recommend the reproduction of these results first, as scores may ultimately vary between different experimental setups.

\begin{table}[t]
	\centering
	\begin{tabular}{l|c|ccc||ccc}
		& 	& \multicolumn{3}{c||}{\textbf{Validation Set}} & \multicolumn{3}{c}{\textbf{Test Set}} \\
		\textbf{Dataset} & \textbf{Method} & \textbf{R-1} & \textbf{R-2} & \textbf{R-L} & \textbf{R-1} & \textbf{R-2} & \textbf{R-L} \\
		\hline
		\hline
		& lead-3 & $19.06$ &  $5.58$ & $13.21$ & $18.90$ & $5.47$ & $13.04$ \\
		\textbf{MLSUM} & lead-$k$ & $14.93$ & $4.12$ & $11.31$ & $15.08$ & $4.17$ & $11.45$ \\
		& LexRank-ST & $15.78$ & $ 3.36$ & $11.52$ & $16.04$ & $ 3.30$ & $11.55$ \\
		\hline
		& lead-3 & $15.19$ & $ 3.46$ & $ 9.10$ & $15.87$ & $ 3.64$ & $ 9.35$ \\
		\textbf{Klexikon} & lead-$k$ & $28.11$ & $ 5.51$ & $12.43$ & $28.34$ & $ 5.50$ & $12.50$ \\
		& LexRank-ST & $27.23$ & $ 4.63$ & $11.48$ & $27.42$ & $ 4.58$ & $11.55$\\
		\hline
		& lead-3 & $16.72$ & $ 2.80$ & $10.51$ & $16.74$ & $ 2.86$ & $10.53$ \\
		\textbf{LegalSum} & lead-$k$ & $14.34$ & $ 2.27$ & $ 8.78$ & $14.36$ & $ 2.34$ & $ 8.78$ \\
		& LexRank-ST & $21.54$ & $ 6.22$ & $12.97$ & $21.35$ & $ 5.99$ & $12.74$\\
		\hline
		& lead-3 & $ 3.31$ & $ 2.25$ & $ 2.72$ & $ 3.31$ & $ 2.19$ & $ 2.67$ \\
		\textbf{EUR-Lex-Sum} & lead-$k$ & $41.74$ & $17.77$ & $16.04$  & $39.42$ & $17.08$ & $15.52$ \\
		& LexRank-ST & $39.37$ & $15.13$ & $15.26$ & $38.48$ & $15.18$ & $15.19$ \\
	\end{tabular}
	\caption{Baseline results for all datasets with available validation and/or test splits. We report ROUGE F1 scores on the filtered datasets.}
	\label{tab:baselines}
\end{table}

\subsection{Impact of Data Filtering}
\begin{displayquote}
	\textbf{Key Finding 4:} After filtering, scores can drop by more than 20 ROUGE-1 points on the MLSUM test set.
\end{displayquote}
To illustrate the effect of dataset filtering on downstream performance, we further compare results on the two most-affected datasets (MLSUM and MassiveSumm).
Without any additional training, we run all available public models on the validation and test portion of MLSUM, for which we also obtain scores on the original unfiltered sets.
Our findings can be seen in \Cref{tab:filtered}, where one can observe a performance drop in \emph{every model}, even those that were not originally trained on the MLSUM dataset itself (\emph{t5-base}).
By far the worst affected are the two baselines constructed from leading sentences, as well as the mT5-small models by users \emph{mrm8488} and \emph{Shahm}. These four models all achieve unreasonably high ROUGE-2 scores before filtering and see a reduction to about one fifth of the original scores after filtering. Upon inspection, we similarly found that these models were ultimately simply re-generating the first tokens from the input article.
These findings are concerning, as they ultimately question the current state-of-the-art on the MLSUM dataset. 
It further validates the necessity of filtering, given that we can ultimately change the course of evaluation and interpretation of models.
For MLSUM, per our results, the \emph{t5-base} model, trained on a related news dataset and utilizing the largest underlying neural model, seems to perform best on filtered datasets while originally lagging behind even a simple lead-3 baseline. This is particularly interesting, because the underlying model checkpoint used is primarily trained on English texts.

\begin{figure}[t]
	\centering
	\begin{subfigure}{0.49\textwidth}
		\includegraphics[width=1.0\textwidth]{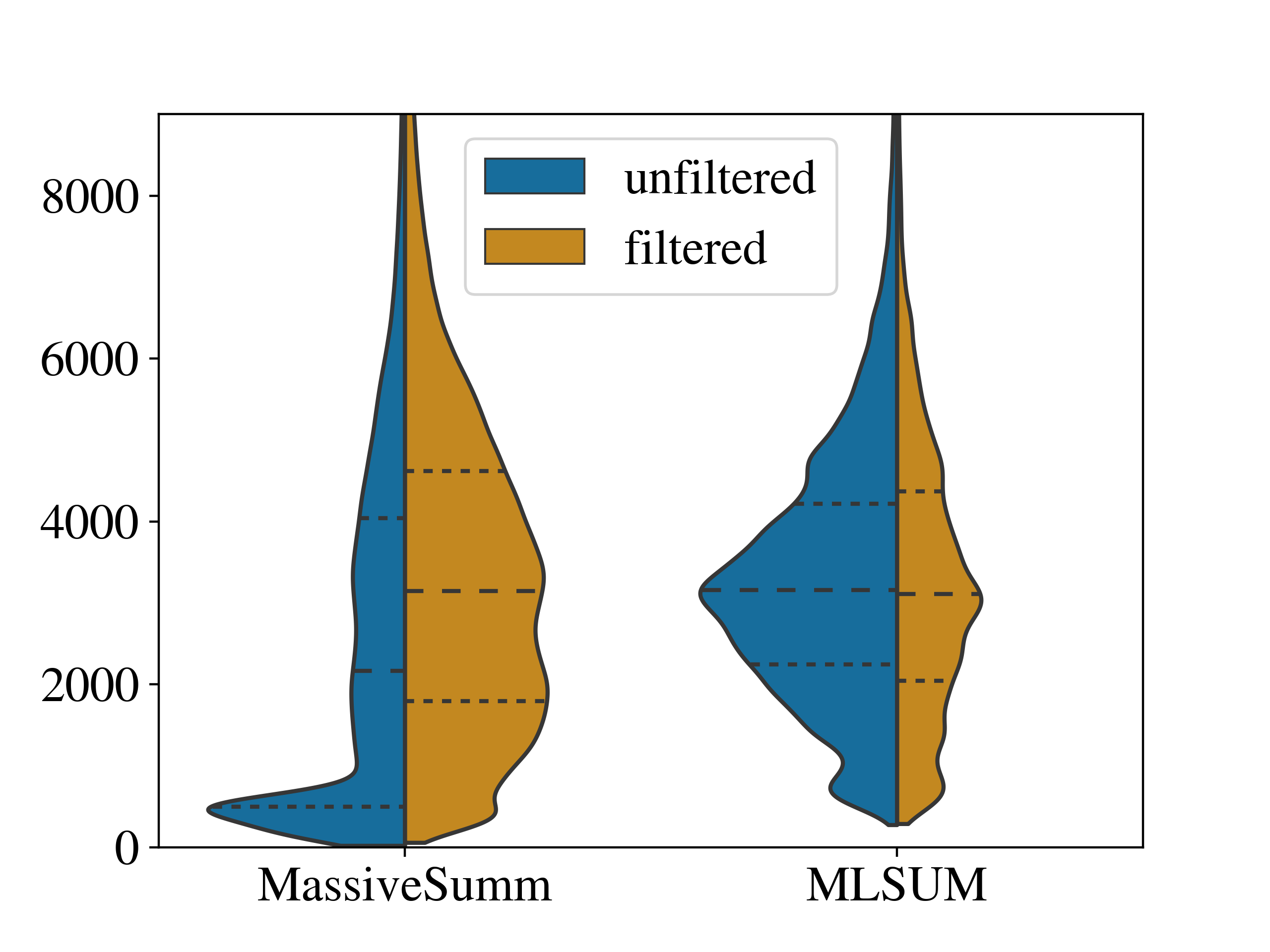}
		\caption{Reference Text Lengths}
	\end{subfigure}%
	\hfill
	\begin{subfigure}{0.49	 \textwidth}
		\includegraphics[width=1.0\textwidth]{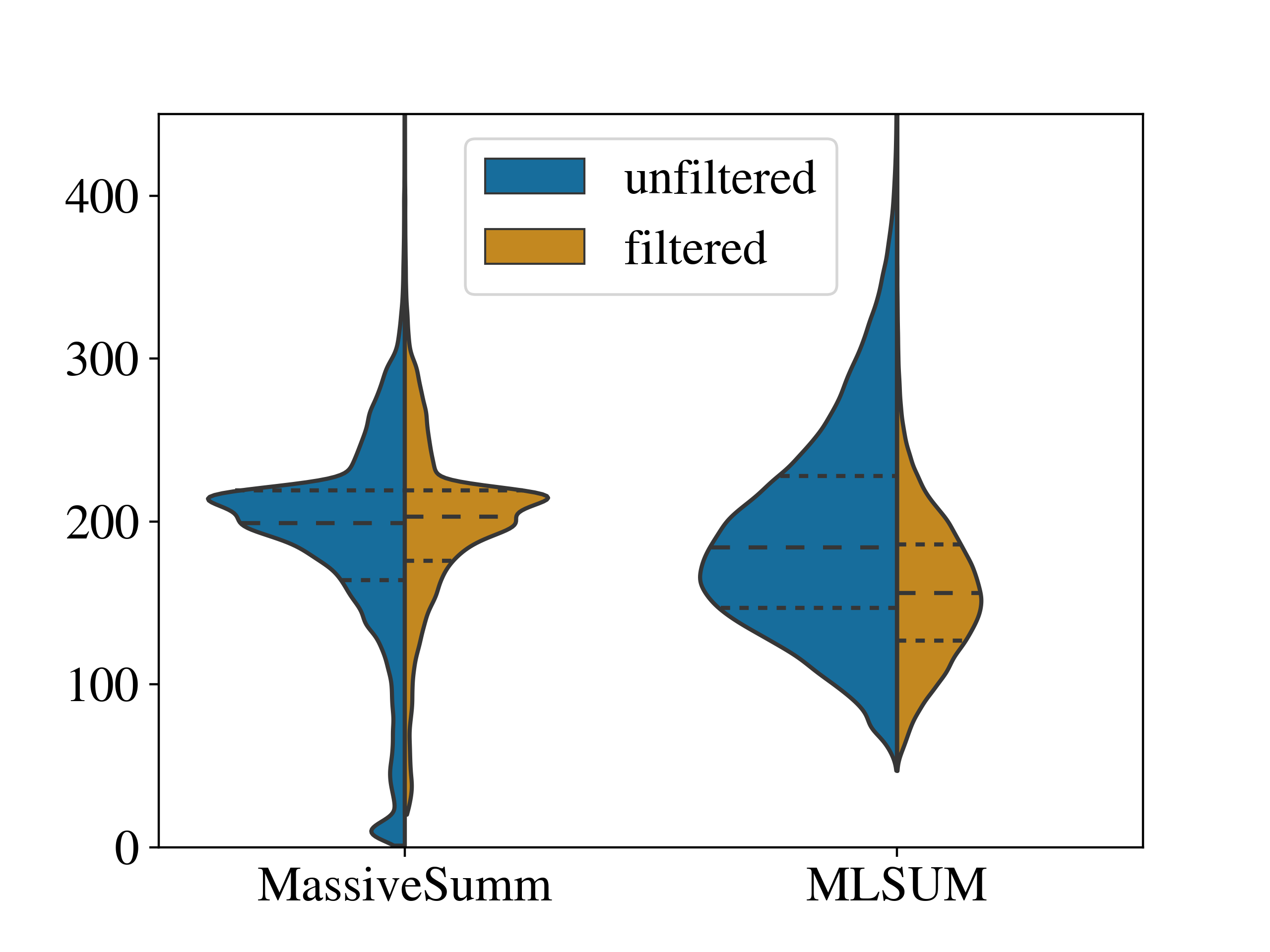}
		\caption{Summary Text Lengths}
	\end{subfigure}
	\caption{Violin plots illustrating the distributional shift on the MLSUM and MassiveSumm training splits through filtering. Black dashed lines indicate mean and quartiles of the distribution.}
	\label{fig:lengths}
\end{figure}

Figure \ref{fig:lengths} further visualizes the impact of filtering on the length distributions of the two heavily affected datasets, MLSUM and MassiveSumm.
Analyzing the resulting changes in more detail, we can observe a more strictly enforced minimum length for both references and summaries in the MLSUM dataset even before filtering. In stark contrast, MassiveSumm is shrunk considerably by the minimum length filter, which in turn shifts the samples towards generally longer reference texts.
Since MLSUM is affected more by the extractiveness filter, one can observe a noticeable change in the mean of the distribution of summary texts, particularly longer ones.\\
Changes in the length distribution, however, do not explain any of the deterioration in raw ROUGE scores; a further indicator that several different evaluation methods need to be combined in order to paint a more complete picture for the realistic performance of models.
We particularly recommend the utilization of violin plots also for the evaluation of system outputs, as they allow the comparison of length estimates by the system in comparison to ground truth data.

%

\subsection{Qualitative Analysis of Generated Summaries}

\begin{displayquote}
	\textbf{Key Finding 5:} With the exception of one work~\cite{aksenov-etal-2020-abstractive}, no \emph{publicly} available system performs experiments beyond simple ROUGE score computation.\\
	\textbf{Key Finding 6:} Despite high reported scores, catastrophic failures can be observed in some systems.\\
	\textbf{Key Finding 7:} All utilized architectures only work with a relatively limited context, proving to be incapable of dealing with long-form summarization.
\end{displayquote}

The first criterion we were looking at when checking for existing systems is the evaluation setting that was used in the respective work.
The findings, reported in \Cref{tab:models}, point towards a more rigorous evaluation setting for models backed by a scientific publication, which comes as no surprise.
However, we also note that these systems are also more likely to withhold their respective models from public access.
This ultimately means that those models can only be judged based on the reported evaluation and no further black-box model checks can be performed on them.
To aggregate the insights gained across these works, most frequently mentioned is the issue of factual consistency~\cite{venzin-etal-2019-swisstext,fecht-etal-2019-swisstext}, which does not bode well for the practical suitability of such systems beyond simple settings.
Secondly, several works also investigate system outputs' fluency~\cite{fecht-etal-2019-swisstext,aksenov-etal-2020-abstractive}, where abstractive models could provide sensible improvements over extractive systems. However, especially for earlier works, consistent generations from language models still prove to be difficult.

\begin{table}[t]
	\hspace*{-0.45cm}
	\setlength{\tabcolsep}{3.5pt}
	\centering
	\begin{tabular}{l|ccc|ccc||ccc|ccc}
		& \multicolumn{6}{c||}{\textbf{MLSUM Validation Split}} & \multicolumn{6}{c}{\textbf{MLSUM Test Split}} \\
		\hline
		& \multicolumn{3}{c|}{\textbf{Unfiltered}} & \multicolumn{3}{c||}{\textbf{Filtered}} & \multicolumn{3}{c|}{\textbf{Unfiltered}} & \multicolumn{3}{c}{\textbf{Filtered}} \\
		\textbf{Model} & \textbf{R-1} & \textbf{R-2} & \textbf{R-L} & \textbf{R-1} & \textbf{R-2} & \textbf{R-L} & \textbf{R-1} & \textbf{R-2} & \textbf{R-L} & \textbf{R-1} & \textbf{R-2} & \textbf{R-L} \\
		\hline
		\hline
		\textbf{Lead-3} & $36.22$ & $26.24$ & $31.89$ & $19.06$ &  $5.58$ & $13.21$ & $37.15$ & $27.48$ & $32.94$ & $18.90$ & $5.47$ & $13.04$ \\
		\textbf{Lead-$k$} & $29.25$ & $20.92$ & $26.51$ & $14.93$ & $4.12$ & $11.31$ & $31.35$ & $22.86$ & $28.58$ & $15.08$ & $4.17$ & $11.45$\\
		\textbf{LexRank-ST} & $18.62$ & $ 6.46$ & $14.26$ & $15.78$ & $ 3.36$ & $11.52$ & $18.83$ & $ 6.45$ & $14.36$ & $16.04$ & $ 3.30$ & $11.55$ \\
		\hline
		\textbf{mrm8488} & $\mathbf{42.77}$ & $31.89$ & $\mathbf{38.93}$ & $21.63$ & $ 6.64$ & $16.32$ & $\mathbf{44.05}$ & $33.44$ & $\mathbf{40.36}$& $21.31$ & $ 6.36$ & $16.09$ \\
		\textbf{ml6team} & $28.17$ & $18.81$ & $26.05$ & $17.08$ & $ 5.03$ & $14.18$ & $28.51$ & $19.52$ & $26.53$ & $16.56$ & $ 4.80$ & $13.78$ \\
		\textbf{T-Systems} &$23.74$ & $11.08$ & $20.34$ & $19.87$ & $ 6.49$ & $16.40$ & $23.67$ & $11.21$ & $20.36$ & $19.20$ & $ 6.11$ & $15.84$ \\
		\textbf{Shahm} & $42.59$ & $\mathbf{31.96}$ & $38.70$ & $21.50$ & $ 6.87$ & $16.15$ & $43.92$ & $\mathbf{33.62}$ & $40.09$ & $21.20$ & $ 6.62$ & $15.79$ \\
		\textbf{t5-base} & $27.54$ & $11.31$ & $20.88$ & $\mathbf{23.31}$ & $ \mathbf{7.19}$ & $\mathbf{16.99}$ & $27.99$ & $11.65$ & $21.20$ & $\mathbf{23.40}$ & $\mathbf{7.20}$ & $\mathbf{16.91}$ \\
	\end{tabular}
	\caption{ROUGE F1 scores on the MLSUM validation and test splits, comparing results with and without data filtering. Across all tested models, a stark drop in performance can be observed. We highlight the highest score for each split in bold.}
	\label{tab:filtered}
\end{table}

To follow our own advice, we manually investigated instances of generated outputs from systems in Table \ref{tab:filtered}.
In addition to samples from the MLSUM dataset, we further tested with instances from the Klexikon and WikiLingua datasets to check for domain generalization.
As others have noted, the factual consistency of abstractive systems is questionable at best, but understated just how badly summaries can deviate from the original.
Several times a reversed order of aggressors and victims (respectively, winners or losers in sports game) was generated, and in one particular instance the context was altered from ``live-saving'' to ``drowning (someone)'' by the summarization system. This happened on ``in-domain samples'' from the MLSUM test set.\\
A similar observation can be made for the syntactic quality of generations, where overfitting of systems becomes particularly apparent during the zero-shot evaluation on other datasets. While it can be expected that the quality of a generated summary may lack in content accuracy or truthfulness, oftentimes no coherent sentence was provided.
Less tragic, but difficult for system comparison, is the multitude of parameters for generation functions. While self-reported scores of public models generally rely on greedily decoded summaries, one model frequently started repeating short sequences of about three words indefinitely until the maximum generation length was reached.
Importantly, such repetitions are not obvious from looking at a ROUGE-based evaluation of model outputs alone, but could be easily suppressed by enabling $n$-gram-based filtering during the generation.\\
We were also able to verify that the highly-scoring models by users \emph{mrm8488} and \emph{Shahm} indeed only copy the leading tokens from the input samples, likely due to training on unfiltered MLSUM splits.
This spells further trouble for ``state-of-the-art'' models, as it requires a deeper examination for determining which summaries are actually better than simple string selection approaches, such as lead-3.
We hypothesize that the same concept used in our \emph{extractiveness} filter can also be applied to generated outputs; with a slightly altered similarity scoring mechanism, e.g., the longest common subsequence algorithm, even near duplicates could be detected and flagged for manual review.

Most prominently though, due to architectural constraints of the underlying neural models, none of the currently public systems is able to capture an input context beyond 512 subword tokens, the default length limit for the Transformer architectures~\cite{vaswani-etal-2017-attention}. In the instance of domain-specific datasets, such as EUR-Lex-Sum, this means that even the length of summary texts exceeds the limitation of models, effectively rendering them useless in this particular context.




\section{Conclusion and Future Work}

Studying the current landscape of German abstractive summarization initially paints a grim picture:
While the general willingness and ease of sharing systems has greatly increased over the past years, around half of the currently known German summarization systems still remain inaccessible to the public.
Of those that \emph{are} available for public scrutiny, a prominent focus on news summarization is still persisting, preventing more broader applications.
Even worse, the most prominent dataset contains severe flaws in the sample quality, leading to models whose generalization capabilities, even in-domain, are severely hampered by the unfiltered data.
This also hints at the general level of care practitioners take with respect to exploratory data analysis, given that several issues can be spotted by simply inspecting just a few samples.
And finally, even models that take care of filtering some of these issues, a qualitative analysis of generations can still reveal catastrophic problems that prevent an ethically responsible deployment of the solution in practice.

However, there are some silver linings at the horizon.
Many of the major data-centric issues can be easily fixed with the introduced quality checks, which can be applied cost-effectively across multiple datasets, as we have demonstrated in this work.
Through publishing our pre-processing pipeline, we hope to encourage others in taking a more data-centric exploration before starting with the ultimate model training.\\
Within just two years, we have also seen an unbelievable influx of available summarization datasets for German, importantly extending past the narrow domains into application-specific fields, such as law and medicine, and totaling more than 700{.}000 samples across publicly available resources.
This hopefully paves the way towards a more consistent and generalized approach in German abstractive summarization research;
should the efforts of the community keep at the current rate, we will likely see meaningful progress within the next year.
The latest trends in the English summarization community also indicate a shift towards greater awareness of long-form summarization~\cite{phang-etal-2022-investigating}; while dedicated long context German (or multilingual) model checkpoints are still absent, we estimate that such systems will become available shortly, serving as a compute-intensive way to escape the current restrictions on input length.

As for our own efforts, we are currently investigating how systems can be designed to work well across multiple domains at once, without the need for several distinct models. This requires careful analysis of the underlying data, as well as a more agnostic training framework to prevent overfitting towards a particular style.

\bibliography{custom} 
\end{document}